\documentclass{article}

\usepackage[backend=biber,maxbibnames=5,style=numeric,maxcitenames=2,backref=false,uniquelist=false,uniquename=false,sorting=none,defernumbers=true, url=false,doi=false,isbn=false]{biblatex}
\usepackage{arxiv}          
\usepackage[utf8]{inputenc} 
\usepackage[T1]{fontenc}    
\usepackage{hyperref}       
\usepackage{url}            
\usepackage{booktabs}       
\usepackage{amsfonts}       
\usepackage{nicefrac}       
\usepackage{microtype}      
\usepackage{mathrsfs}       
\usepackage{mathtools}      
\usepackage{multicol}       
\usepackage{multirow}       
\usepackage{xspace}         
\usepackage{wrapfig}        
\usepackage{subcaption}     
\usepackage{float}          
\usepackage[labelfont=bf,justification=raggedright,singlelinecheck=false]{caption}  
\usepackage{siunitx}        
\newcommand{\sn}[2]{{#1} \times 10^{#2}} 

\DefineBibliographyStrings{american}{phdthesis = {PhD Dissertation}}
\renewbibmacro{in:}{%
	\ifentrytype{article}{}{%
		\printtext{\bibstring{in}\intitlepunct}}}
\renewbibmacro*{volume+number+eid}{%
	\printfield{volume}%
	\setunit*{\addnbspace}
	\printfield{number}%
	\setunit{\addcomma\space}%
	\printfield{eid}}
\renewbibmacro*{volume+number+eid}{%
	\setunit*{\addcomma\space}
	\printfield{volume}%
	\setunit*{\addcomma\space}
	\printfield{number}%
	\setunit{\addcomma\space}%
	\printfield{eid}
}
\DeclareFieldFormat[article]{volume}{\bibstring{volume}~#1}
\DeclareFieldFormat[article]{number}{\bibstring{number}~#1}
\newbibmacro{string+doiurlisbn}[1]{%
	\iffieldundef{doi}{%
		\iffieldundef{url}{%
			\iffieldundef{isbn}{%
				\iffieldundef{issn}{%
					#1%
				}{%
					\href{http://books.google.com/books?vid=ISSN\thefield{issn}}{#1}%
				}%
			}{%
				\href{http://books.google.com/books?vid=ISBN\thefield{isbn}}{#1}%
			}%
		}{%
			\href{\thefield{url}}{#1}%
		}%
	}{%
		\href{http://dx.doi.org/\thefield{doi}}{#1}%
	}%
}
\DeclareFieldFormat{title}{\usebibmacro{string+doiurlisbn}{\mkbibemph{#1}}}
\DeclareFieldFormat[article,incollection]{title}%
{\usebibmacro{string+doiurlisbn}{\mkbibquote{#1}}}

\title{Univariate Long-term Municipal Water Demand Forecasting}

\author{
  Blake VanBerlo\\
    VanBerlo Consulting\\
    London, Canada \\
    \texttt{blake@vanberloconsulting.com} \\
  \And
  Matthew A. S. Ross \\
   Artificial Intelligence\\
   Information and Technology Services\\
   The Corporation of the City of London\\
   London, Canada \\
   \texttt{maross@london.ca} \\
  \And
  Daniel Hsia \\
   Water Demand Management\\
   Environmental and Engineering Services\\
   The Corporation of the City of London\\
    London, Canada \\
    \texttt{dhsia@london.ca} \\
}

\begin{document}
\maketitle

\begin{abstract}
This study describes an investigation into the modelling of citywide water consumption in London, Canada. Multiple modelling techniques were evaluated for the task of univariate time series forecasting with water consumption, including linear regression, Facebook's Prophet method, recurrent neural networks, and convolutional neural networks. Prophet was identified as the model of choice, having achieved a mean absolute percentage error of $2.51\%$, averaged across a 5-fold cross validation. Prophet was also found to have other advantages deemed valuable to water demand management stakeholders, including inherent interpretability and graceful handling of missing data. The implementation for the methods described in this paper has been open sourced, as they may be adaptable by other municipalities.
\end{abstract}


\section{Introduction}
\label{sec:intro}

Forecasting is a commonplace task in municipal government. City departments have an obligation to their citizens to limit their expenditures on investments that promote the long-term stewardship of public resources. Municipal water demand is a typical example of a quantity that would be useful to model for forecasting purposes to enable better budgeting and infrastructure planning. The work described in this paper is an application of machine learning forecasting methods to accurately model and predict water consumption in the city of London, Canada.

A reasonably correct forecast of aggregate water demand may serve a collection of purposes in municipal government. A major use for forecasts of water consumption is to produce revenue projections to be incorporated into municipal budgets. Such forecasts, if segmented by rate class and extended over a few years, can form the basis of revenue projection from the utility. A reasonably accurate forecast may be used to identify and justify major investments in water infrastructure. Further, a model for water consumption may provide insight into repeating and/or overarching patterns that may exist in consumption. For instance, some forecasting models entail characterization of seasonal variation of water demand. The context-specific knowledge that may emerge from forecasting places a valuable emphasis on model interpretability; that is, models whose predictions may be understood based on knowledge of the model itself. Lastly, forecasting methods may be applied to various strata of the population, enabling comparisons of the habits of each group and derivation of useful customer behaviour insights.

This paper details the search for and evaluation of a univariate model for long-term water consumption forecasting in London, Canada. The Prophet modelling algorithm~\cite{Taylor2018} was ultimately selected to produce the city's new long-term water demand forecasting model. Section~\ref{sec:related-works} provides a brief survey of related work in water demand forecasting. The procedure that was followed to arrive at the final model will then be presented in Section~\ref{sec:methods}, along with its justification. The evaluation of the candidate models will be detailed in Section~\ref{sec:results}, along with forecasting and interpretability details pertaining to the final model. Finally, Section~\ref{sec:discussion} will explore the results and propose directions for future work.
 
\section{Related Work}
\label{sec:related-works}

The use of modern machine learning methods to forecast utility demand is hardly a novel concept, and is currently being explored on several fronts in various locations. For instance, researchers at the University of Western Ontario applied support vector regression to forecast household electricity consumption~\cite{Zhang2019}. More recently, researchers at the same university investigated an evolutionary artificial neural network (ANN) approach to forecast residential electrical load demand~\cite{Gomez-rosero2020}. ANN models have also be applied to predict natural gas consumption in Szczecin, Poland~\cite{Szoplik2015} and in multiple Greek cities~\cite{Anagnostis2020}. Hereafter, we focus our discussion on efforts to forecast water demand.

Forecasts of water consumption typically fall into one of two categories: short-term and long-term. Both involve predicting some sort of indicator of future water consumption, but are often based on different sets of features and underlying data. Hanyu Liu's master's thesis focuses on development of short- and long-term forecasting models for water demand in Edmonton, Canada~\cite{Liu2020}. Liu asserts that, in the short-term scenario, meteorological data and recent consumption data are important predictors of water consumption~\cite{Liu2020}. In his short-term modelling, Liu employs ANN approaches to learn daily and weekly forecasting models from a variety of predictors, including water consumption, temperatures, precipitation, and date characteristics (e.g. day of week) over the days/weeks preceding the forecast. ~\textcite{Adamowski2012} investigated the use of discrete wavelet transforms and ANNs for predicting short-term water consumption from weather and recently observed water demand, achieving a relative root mean square error of $1.59\%$. \textcite{Shah2018} studied the application of an ANN known as recurrent neural networks (RNN) to short-term forecasting in Indiana, United States. Training their models using weather and population data, their team found that RNN models outperformed multiple linear regression and feed-forward neural networks, achieving a test set error of approximately $2.5\%$~\cite{Shah2018}. Despite their prevalence in the literature, ANNs are not the only model used to forecast short-term demand. The auto-regressive integrated moving average (ARIMA) method has previously been applied to hourly water demand forecasting~\cite{ShvartserLeonid1993FHWD}. A 2020 study by \textcite{Du2020} proposes a method by which ARIMA is used to produce a short-term daily forecast, which is then modified by a Markov model to account for future trend.

Our work seeks to address the problem of long-term water demand forecasting. A $4$-year forecast length was selected, so as to align with stakeholder multi-year budget projection obligations. In the long-term scenario, several predictors useful for short-term modelling are no longer useful. For instance, the volatility of weather data and its absence for future years renders it ineffective for forecasting on the scale of years into the future. Multiple models have been proposed that take advantage of different predictors. System dynamics (SD) modelling has been used to forecast long-term water demand in the past. For instance, \textcite{QiCheng2011Sdmf} constructed a SD model from macroeconomic predictors to predict residential water usage. \textcite{WangKai2018Mwpa} developed a SD model based on consumer uses of water that predicted weekly water consumption in Calgary, Alberta through 2040. Multiple models were created to account for possible variation in water policy~\cite{WangKai2018Mwpa}. Validation of the model for the years 2005-2015 revealed a relative root mean squared error (RMSE) of $5.76\%$. Similar to the short-term scenario, ANNs frequently appear in the literature for long-term modelling solutions. \textcite{Shrestha2020} trained an ANN model to predict yearly water consumption data based on the number of water connections, water tariff rate, fraction of the population in university, and annual rainfall. Despite their satisfaction with the ANN's performance, the training set was limited to $16$ years. Aside from the small dataset, a major limitation of their approach was the necessity to produce forecasts for the predictors in order to generate long-term water demand forecasts. As a result, uncertainty in the forecasts of the predictors may compound in the ANN's long-term forecasts. \textcite{Liu2020} addresses this shortcoming in his thesis, where he suggests a hybrid SD-ANN model for long-term weekly municipal water demand forecasting in Edmonton, Alberta. His approach considers multiple long-term climate scenarios and produces separate models for each~\cite{Liu2020}. The SD component models indoor water consumption, modelling several domestic uses for water, while outdoor water consumption is predicted by an ANN from temperature and precipitation data computed under varying climate scenarios~\cite{Liu2020}.  He reports a mean absolute error (MAE) of \SI{87}{ML} for an average demand of \SI{2471}{ML} per timestep~\cite{Liu2020}. The long-term forecasting efforts described in this section tend to rely on assumptions about the future values of predictors, and tend to have a timestep of $1$ week or greater.

\section{Methods}
\label{sec:methods}

\subsection{Data Preprocessing}
\label{subsec:data-preprocessing}

As demonstrated by previous work in long-term water demand forecasting, most models create forecasts for a duration of at least a week. We chose a daily timestep, hoping to produce as detailed forecasts as possible. In London, each client's total consumption of water is recorded for each billing period. A billing period is approximately one month in duration, with variance of a few days resulting from the method of data collection from the physical water meter. Additionally, billing periods are offset between clients since meter reads do not take place all at the same time, nor will they be identical month-to-month across the same customer. A method was devised to calculate an estimation for daily citywide water consumption. By producing a daily estimate, we increased the resolution of our data from an (approximate and staggered) monthly time step to a daily time step, resulting in a significantly greater total number of time series records in the dataset. A single client's water consumption for a particular date, $d$, was estimated by dividing their total consumption, $C_{client}$, over the length of the client's billing period (in days) in which $d$ falls in, with inclusive start and end dates $b_{start}$, $b_{end}$, where $d \in [b_{start}, b_{end}]$. The city-wide consumption on date $d$, $C_{d_{total}}$, is then the sum of all clients' estimated daily consumption on date $d$ (Equation \ref{eq:cons-estimate}). This estimation method assumes that clients consume water uniformly over the duration of a billing period. Of course, this assumption does not likely hold true --- water consumption fluctuates over the short term and depends on dynamic weather phenomena~\cite{Liu2020}. The day of the week may also have a large impact on water demand, as water demand can vary on weekdays versus weekends. We argue that this assumption is appropriate for the problem of long-term forecasting, as we are not concerned with short-term irregularities in consumption. Rather, we value forecasts that highlight overall future patterns and trend. Although the estimated daily consumption may appear unduly smooth, the total water consumption over billing periods is preserved in this estimation. Note that even if we determined a monthly estimate for citywide demand, the uniform consumption assumption would still be inevitable, since clients' billing periods were staggered in the raw dataset.

\begin{equation}
\label{eq:cons-estimate}
    C_{d_{total}} = \sum_{clients} \frac{C_{client}}{b_2 - b_1 + 1}
\end{equation}

\begin{figure}
    \centering
    \includegraphics[width=\linewidth]{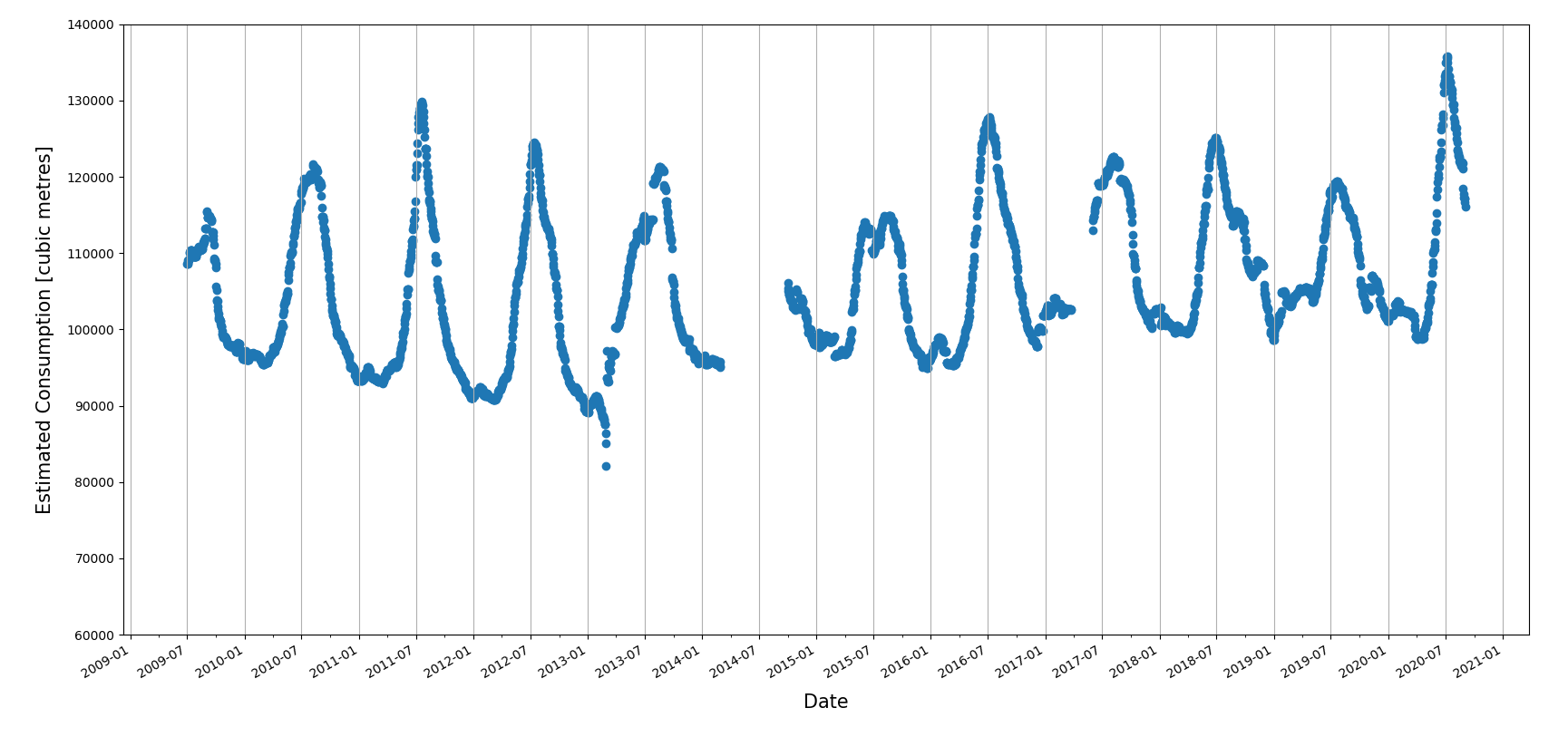}
    \caption{Citywide daily water consumption estimates (in cubic metres) computed from water billing data.}
    \label{fig:preprocessed-data} 
\end{figure}

The consumption estimates produced for each date available in the raw dataset comprised the preprocessed dataset. For the purposes of this investigation, daily estimated water consumption was the only feature in the preprocessed dataset, rendering this a univariate forecasting problem. Estimates were computed for dates between July 7, 2009 and September 2, 2020 inclusive. Unfortunately, billing data could not be recovered for particular date ranges in 2014 and 2017. Consequently, we were unable to obtain consumption estimates for the inclusive date ranges of March 1, 2014 to September 30, 2014, and for March 25, 2017 to May 31, 2017. Figure \ref{fig:preprocessed-data} depicts the daily consumption estimates that were computed from the available water billing data.

\subsection{Candidate Models}
\label{subsec:candidate-models}

As noted in section~\ref{sec:related-works}, several modelling approaches to water demand forecasting have been proposed. We limit our investigation to four candidate models. Given the preponderance of ANNs in the literature, we selected two different ANN architectures with varying mechanisms for handling time series data. To establish a baseline, we considered an ordinary least squares (OLS) regression model, which was the preexisting method used for water demand forecasting in London. Finally, Facebook's Prophet method~\cite{Taylor2018} was included as an intuitive alternative to ANN models.

Two ANN models were considered, each utilizing different architectural components useful for handling time series data. First, a RNN model was investigated. Input to the model consisted of water consumption estimates over $T_x$ consecutive days (where $T_x$ is the input sequence length), and the output was the predicted consumption for the following day. The model consisted of a long short-term memory (LSTM) layer, followed by $2$ fully connected layers with rectified linear activation, and a single-node output layer with linear activation. The second ANN had the same input/output formulations. Its architecture consisted of at least one 1-dimensional convolutional neural network (1D-CNN) layers, followed by $2$ fully connected layers with rectified linear activation, and a single-node output layer with linear activation. Hereafter, these two model architectures are referred to as the LSTM-RNN model and the 1D-CNN model respectively. Both networks make predictions for a particular day based on daily consumption from the last $T_x$ days. Each was trained using the Adam stochastic optimization technique~\cite{KingmaDiederikP2014AAMf}. When training the ANN models, the most recent records in the training set were detached to form a validation set, which informed early stopping, a technique employed to deter overfitting~\cite{Prechelt1998}.

Lastly, Prophet is a forecasting method proposed by Facebook researchers that consists of a combination of configurable functions fitted to time series data~\cite{Taylor2018}. It is composed of smaller components whose behaviours may be in part dictated by developers, lending itself to what the method's creators claim is an "analyst-in-the-loop" approach to forecasting. Prophet is a composite function, consisting of components for piecewise linear or logistic trend, periodic effects (annual, weekly, or custom periods), and holiday influences. Based on input by domain experts, Prophet can be configured to accommodate different formulations of time series problems. In the case of long-term water demand, piecewise linear trend was preferred to logistic trend because a carrying capacity for the aggregate water demand in London is unknown. Prophet models a piecewise linear trend by identifying changepoints in time series data (i.e. dates at which the slope of the linear trend changes) and produces linear trends for the data between changepoints~\cite{Taylor2018}. As evident in Figure~\ref{fig:preprocessed-data}, water consumption in London has an annual seasonality. Prophet calculates a Fourier series to approximate a periodic function that captures seasonal effects applied to the underlying trend~\cite{Taylor2018}. Finally, unique parameters for holidays can be fitted depending on the locale~\cite{Taylor2018}. Equation~\ref{eq:prophet}, taken from the original Prophet publication, summarizes the simple relationship between Prophet's forecast and its constituent functions~\cite{Taylor2018}.

\begin{equation}
\label{eq:prophet}
    y(d) = g(d) + s(d) + h(d) + \epsilon_d
\end{equation}

In the above equation, $y(d)$ is Prophet's prediction for date $d$, $g(d)$ is the trend component, $s(d)$ is the seasonality component, $h(d)$ is the holiday component, and $\epsilon_d$ is a normally distributed error term~\cite{Taylor2018}. As a result, we specified that Prophet include a periodic annual function to capture seasonal effects. Although the daily consumption estimates are fairly smooth as a result of our daily consumption estimate methodology, we also included a periodic weekly function in the model. Finally, we identified all provincial and national holidays impacting work schedules. Prophet estimates individual holidays' effects and carries them forward in future forecasts. The result was a Prophet model with linear trend, entailing yearly and weekly seasonality, along with holiday effects.

\subsection{Hyperparameter Optimization}
\label{subsec:hyperparameter-optimization}

\begin{figure}
    \centering
    \includegraphics[width=0.7\linewidth]{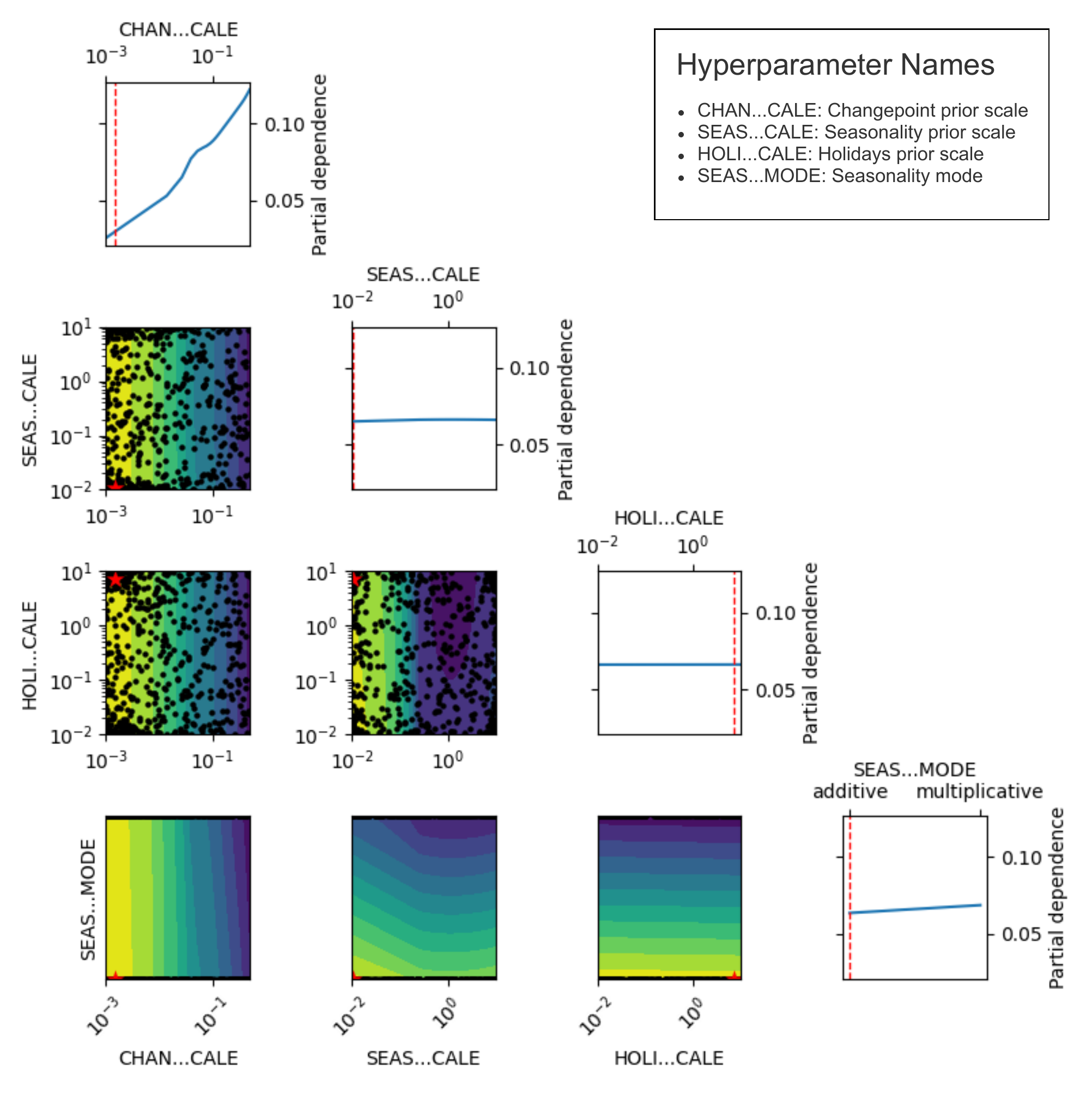}
    \caption{Partial dependence plots (PDP) for average test set MAPE during cross validation (the objective function) for Prophet, created from the results of a Bayesian hyperparameter optimization. The coloured plots are two-dimensional PDPs. Brighter regions correspond to lower values for the objective. Black dots correspond to values of hyperparameters for each iteration of the optimization, and the red star indicates the best iteration. The graphs on the diagonal are one-dimensional PDPs for each hyperparameter considered. The red line indicates the best iteration. Final hyperparameter values were decided based on which values for the hyperparameters corresponded to lower values of the objective.}
    \label{fig:hparam-pdp} 
\end{figure}

Prior to evaluating the four proposed models, hyperparameter optimization was conducted to yield information on the influence of a variety of hyperparameters. We chose to use Bayesian optimization based on Gaussian processes to efficiently guide testing of various hyperparameter values~\cite{Louppe2016}. Each model was given $500$ iterations for the optimization. During each iteration, a $5$-fold cross validation is conducted using a combination of hyperparameters specific to that iteration. Section~\ref{subsec:model-selection-criteria} provides specifics about the time series cross validation technique that was used. The mean absolute percentage error (MAPE) on the test set averaged across the $5$ folds was used to compare combinations of hyperparameters from different iterations. To determine the optimal hyperparameters for a particular model, the list of combinations from each run was examined, along with the corresponding partial dependence plots (see Figure~\ref{fig:hparam-pdp} for an example). Final hyperparameter values and the ranges from which they were selected are presented in Table~\ref{tab:hparam-vals}.

\begin{table}
  \centering
  \small{
  \begin{tabular}{c c c c c c c}
    \toprule
    \textbf{Model}                & \textbf{Hyperparameter} & \textbf{Range type} & \textbf{Range} & \textbf{Determined value} \\
    \toprule
    \multirow{1}{*}{OLS}          & Input sequence length ($T_x$)     & Uniform integer      & $[30, 365]$              & $300$ \\
    \midrule
    \multirow{9}{*}{LSTM-RNN}     & Input sequence length ($T_x$)     & Uniform integer      & $[30, 365]$              & $200$ \\
                                  & Batch size                        & Set                  & ${16, 32}$               & $32$ \\
                                  & Patience (early stopping)         & Uniform integer      & $[5, 15]$                & $15$ \\
                                  & Learning rate                     & Logarithmic float    & $[10^{-5}, 10^{-3}]$     & $10^{-5}$ \\
                                  & Loss function                     & Set                  & \{MSE, MAE, Huber\}      & MSE \\
                                  & Units in LSTM layer               & Set                  & $\{8, 16, 32, 64, 128\}$   & $4$ \\
                                  & Units in 1\textsuperscript{st} fully connected layer               
                                                                      & Set                  & $[32, 64, 128]$          & $64$ \\
                                  & Units in 2\textsuperscript{nd} fully connected layer               
                                                                      & Set                  & $[16, 32, 64]$           & $64$ \\
                                  & Dropout rate                      & Uniform float        & $[0.0, 0.5]$             & $0$ \\
    \midrule
    \multirow{11}{*}{1D-CNN}  & Input sequence length ($T_x$)    & Uniform integer      & $[30, 365]$              & $180$ \\
                                  & Batch size                        & Set                  & ${16, 32}$               & $32$ \\
                                  & Patience (early stopping)         & Uniform integer      & $[5, 15]$                &  $5$ \\
                                  & Learning rate                     & Logarithmic float    & $[10^{-5}, 10^{-3}]$      & $\sn{3}{-4}$ \\
                                  & Loss function                     & Set                  & \{MSE, MAE, Huber\}      & MAE \\
                                  & Filters in 1\textsuperscript{st} 1D convolutional layer                      
                                                                      & Set                  & $\{8, 16, 32\}$            & $16$ \\
                                  & Convolutional kernel size         & Set                  & $[3, 5]$                 & $3$ \\
                                  & Convolutional stride              & Uniform integer      & $[1, 2]$                 & $1$ \\
                                  & Numbers of 1D convolutional layers 
                                                                      & Uniform integer      & $[1, 4]$                 & $2$ \\
                                  & Units in 1\textsuperscript{st} fully connected layer               
                                                                      & Set                  & $[32, 64, 128]$          & $64$ \\
                                  & Units in 2\textsuperscript{nd} fully connected layer               
                                                                      & Set                  & $[16, 32, 64]$           & $32$ \\
                                  & Dropout rate                      & Uniform float        & $[0, 0.5]$               & $0$ \\
    \midrule
    \multirow{4}{*}{Prophet}      & Changepoint prior scale           & Logarithmic float      & $[0.001, 0.5]$         & $0.001$ \\
                                  & Seasonality prior scale           & Logarithmic float      & $[0.01, 10]$           & $0.01$ \\
                                  & Holidays prior scale              & Logarithmic float      & $[0.01, 10]$           & $0.01$ \\
                                  & Seasonality mode                  & Set     & \{additive, multiplicative\}          & additive \\
    \bottomrule
  \end{tabular}}
  \vspace{8pt}
  \caption{Ranges and final values of hyperparameters of each model}
  \label{tab:hparam-vals}
\end{table}

\subsection{Model Evaluation}
\label{subsec:model-selection-criteria}

\begin{figure}
    \centering
    \includegraphics[width=0.7\linewidth]{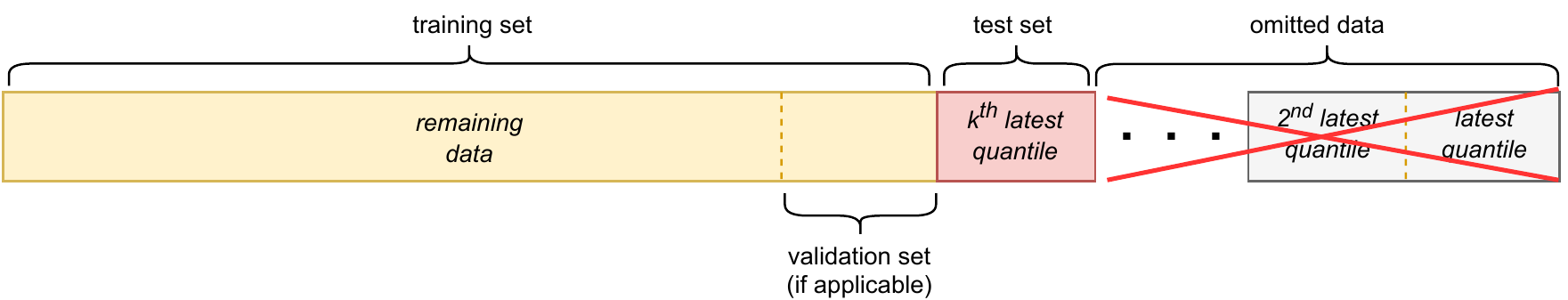}
    \caption{The cross validation strategy, inspired by the rolling origin method~\cite{TashmanLeonardJ2000Otof}. For the $k^{th}$ training experiment, the $(k + 1)^{th}$ quantile of consumption data comprises the test set. All preceding data becomes the training set, unless the model is a neural network; in which case, the latest $10\%$ of the training data is allocated to the validation set. The latest $k$ quantiles are excluded from the experiment.}
    \label{fig:cross-validation} 
\end{figure}

After having selected the final hyperparameter values for each of the four models in consideration, model evaluation was conducted. The procedure for model evaluation was twofold. First, a 5-fold cross validation was completed for each model. Since we are working with time series data, cross validation similar to the rolling origin methodology proposed by \textcite{TashmanLeonardJ2000Otof} was employed, whereby different models are trained using progressively smaller fractions of the training set and evaluated on the data immediately following the curtailed training set for a particular fold. For our purposes, $4$ models were trained using the $4$ most recent sextiles of preprocessed data as the test sets for each fold, and the training sets consist of all data preceding the test set. Figure~\ref{fig:cross-validation} portrays the cross validation strategy that was employed. This strategy ensures that no individual model performs well due to a fortunate quirk in the most recent consumption data. Assessing multiple trained models with different test sets starting during various seasons provides confidence in the modelling technique in question. Since this study was completed during the COVID-19 pandemic, this cross validation strategy is especially crucial. A glance at the preprocessed data (Figure~\ref{fig:preprocessed-data} demonstrates the peaked irregularity in water consumption through the Spring and Summer months of 2020. It would be unfair to evaluate the model only on its performance for a test set spanning the COVID-19 pandemic. The primary metric of interest was mean absolute percentage error (MAPE) on the test set, but mean squared error (MSE) and root mean squared error (RMSE) were also considered. Since recurrent neural architectures do not handle missing data well, it was decided that, at this stage, models be evaluated on the largest and most recent continuous date range for which we had consumption data. Recall from section~\ref{subsec:data-preprocessing} that there were $2$ ranges of missing data. Therefore, the models were trained on all consumption data between June 1, 2017 and September 2, 2020, which comes to $1190$ days in total.

Aside from the objective performance of the model, other aspects were included in the evaluation of each model. Factors considered included model interpretability, handling of missing data, and degree of opportunity for domain expertise to improve the forecast. These criteria are explored further in section~\ref{sec:results}, in tandem with the performance metrics.

\subsection{Implementation Details}
\label{subsec:implementation-details}

The code for this study was written in Python 3, and it is available open source in our public GitHub repository\footnote{\url{https://github.com/aildnont/water-forecast}}. Training experiments were conducted using a personal computer running Windows 10 equipped with an Intel\textsuperscript{\textregistered} Core\textsuperscript{TM} i7-8750H CPU at \SI{2.2}{GHz}. ANN training was accelerated through use of a NVIDIA GeForce\textsuperscript{\textregistered} GTX 1050 Ti graphics processing unit with \SI{4}{GB} of memory.

\section{Results}
\label{sec:results}

\subsection{Model Selection}
\label{subsec:model-selection}

Armed with the hyperparameter combinations derived from the Bayesian hyperparameter search discussed in section~\ref{subsec:hyperparameter-optimization}, 5-fold cross validation was employed, using the strategy outlined in section~\ref{subsec:model-selection-criteria}. Test set performance metrics calculated during cross validation include MAE, MAPE, MSE, and RMSE. The metric of primary interest was MAPE. Table~\ref{tab:model-comparison} summarizes the results of 4-fold time series cross validation for each of the four candidate models, displaying values for test set performance metrics averaged over each fold. Prophet achieved the lowest MAPE score overall, which was the primary quantifiable selection criteria. Additionally, Prophet achieved the lowest MAE and MSE scores averaged over cross-validation. Observe that LSTM-RNN performed nearly as well as Prophet. 

\begin{table}
  \centering
  \begin{tabular}{c c c c c c c}
    \toprule
    \textit{Model}      & \multicolumn{4}{c}{\textit{Mean metric value [standard deviation]}} \\
                       & MAE (m\textsuperscript{3}) & MAPE (\%) & MSE (m\textsuperscript{6}) & RMSE (m\textsuperscript{3}) \\
    \midrule
     OLS &              $\sn{5.73\:[3.88]}{3}$  & $5.22\:[3.30]$  & $\sn{6.14\:[6.13]}{3}$ & $\sn{6.79\:[4.51]}{3}$ \\
     LSTM-RNN &         $\sn{4.31\:[2.76]}{3}$ & $3.87\:[2.30]$ & $\sn{3.02\:[3.44]}{7}$ & $\mathbf{\sn{4.83\:[3.02]}{3}}$ \\
     1D-CNN &      $\sn{4.50\:[2.65]}{3}$ & $4.02\:[2.13]$  & $\sn{3.41\:[3.52]}{7}$ & $\sn{5.27\:[2.91]}{3}$ \\
     Prophet &          $\mathbf{\sn{4.30\:[1.57]}{3}}$  & $\mathbf{3.83\:[1.22]}$ & $\mathbf{\sn{2.82\:[1.97]}{7}}$ & $\sn{4.99\:[2.10]}{3}$ \\
    \bottomrule
  \end{tabular}
  \vspace{8pt}
  \caption{Mean and standard deviations for test set performance metrics across the $4$ folds from rolling origin cross validation. All models were trained on the longest continuous recent subset of data (June 1, 2017 to September 2, 2020).}
  \vspace{-10pt}
  \label{tab:model-comparison}
\end{table}

Prophet was designated as the model of choice. Aside from the quantitative selection criteria outlined above, other capabilities of Prophet contributed to its ultimate selection as the best model candidate. Prophet has multiple advantages over neural networks. First, Prophet does not suffer when there are missing values in the dataset~\cite{Taylor2018}. Recall from section~\ref{subsec:data-preprocessing} that major intervals of data were missing for the years 2014 and 2017. Simply deleting missing values introduces potential bias into time series datasets~\cite{Saad2020}. Although we acknowledge the existence of novel time series imputation solutions~\cite{Saad2020}, it was determined by stakeholders that for this scenario, imputation was not an option. A major advantage of Prophet is that it employs a generative model, avoiding the requirement that time series data points be evenly distributed~\cite{Taylor2018}. Prophet therefore enabled us to train on the entire time series dataset, stretching from 2009 to 2020.

A second major advantage of the Prophet model is that it is inherently interpretable by design. ANN models are not inherently interpretable, often requiring post-processing to yield explanations for their predictions.  As this is a municipal government application of machine learning, it is desirable (and arguably necessary) to be able to justify and explain the model's predictions to any inquiring party. Several interpretability techniques exist that can be applied to ANN models~\cite{Molnar2019}; however, they often involve approximations and would require extra effort to explore. As mentioned in section~\ref{subsec:candidate-models}, Prophet is composed of multiple components that, when combined, compose the output of the whole model. The individual components can be visualized and serialized to further end-user understanding of the basis of the model's predictions.

In summary, the superior performance of Prophet as evidenced by comparing cross validation scores, along with its ability to handle missing data and its inherent interpretability, make it the ideal candidate model. As a result, Prophet was chosen as the model to forecast water demand for the City of London.

\subsection{Final Model}
\label{subsec:final-model}

\begin{wraptable}{R}{0.3\linewidth}
    \centering
    \small{
    \vspace{0pt}
    \begin{tabular}{c c}
        \toprule
        \textbf{Hyperparameter}              & \textbf{Value} \\
        \midrule
        Changepoint prior scale     & $0.001$ \\
        Seasonality prior scale     & $0.01$ \\
        Holidays prior scale        & $0.01$ \\
        Seasonality mode            & additive \\
        \bottomrule
    \end{tabular}}
    \caption{Final hyperparameter values for the Prophet model, trained on the complete dataset.}
    \vspace{-5pt}
    \label{tab:final-prophet-hyperparameters}
\end{wraptable}

\begin{figure}
    \centering
    \includegraphics[width=\linewidth]{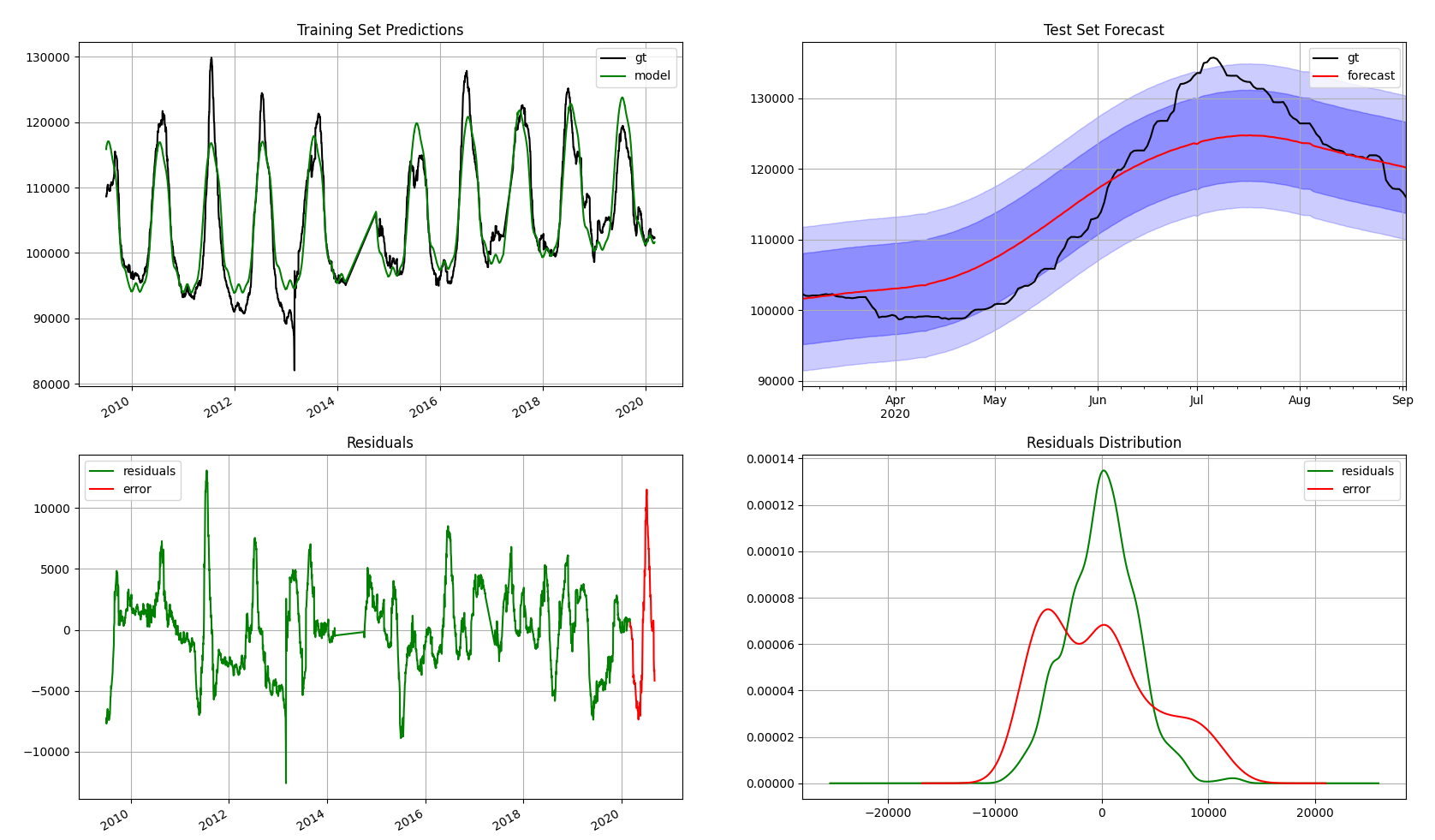}
    \caption{A collection of visualizations evaluating the performance of the final Prophet model. In the top left, the model's predictions on the training set are contrasted with the ground truth daily consumption (i.e. "gt"). The top right graph compares the model's forecast for the 6-month test set with the ground truth consumption from the test set. The bottom left graph depicts residuals (training set MAE) and test set error (MAE). The bottom right graph compares the distribution of residuals to test error. All water consumption values on the y axes are expressed in cubic metres.}
    \label{fig:final-prophet-evaluation} 
\end{figure}

After the decision had been made to select Prophet as the model for water demand forecasting, we were able to train Prophet on the entire available preprocessed dataset, dating back to July 2009. An additional hyperparameter search was conducted to inform final hyperparameter choices for the Prophet model in production. For each combination of hyperparameters, rolling origin 5-fold cross validation was conducted by taking the $5$ folds to be the most recent $5$ deciles of the complete dataset. The final hyperparameters determined for the Prophet model are shown in Table~\ref{tab:final-prophet-hyperparameters}. Using the newly decided hyperparameters, cross validation was executed again, attaining the metrics shown in Table~\ref{tab:final-prophet-cross-validation}. To get a sense of how the final Prophet model fit the data, consider the visualizations in Figure~\ref{fig:final-prophet-evaluation}. For this model, the test set was taken to be the most recent 6 months of data (early March to early September 2020). Despite the test set coinciding with the COVID-19 pandemic, the forecast is remarkably faithful to the ground truth consumption values.

\begin{table}
  \centering
  \begin{tabular}{c c c c c c c}
    \toprule
    \multicolumn{4}{c}{\textit{Mean metric value [standard deviation]}} \\
    MAE (m\textsuperscript{3}) & MAPE (\%) & MSE (m\textsuperscript{6}) & RMSE (m\textsuperscript{3}) \\
    \midrule
    $\sn{2.78}{3}\:[\sn{7.67}{2}]$ & $2.51\:[0.7]$ & $\sn{1.36}{7}\:[\sn{4.93}{6}]$ & $\sn{3.63}{3}\:[\sn{6.65}{2}]$   \\
    \bottomrule
  \end{tabular}
  \vspace{8pt}
  \caption{Mean and standard deviations for test set performance metrics across the $5$ folds from rolling origin cross validation for the Prophet model trained on the complete dataset.}
  \vspace{-10pt}
  \label{tab:final-prophet-cross-validation}
\end{table}

\begin{figure}
    \centering
    \includegraphics[width=0.8\linewidth]{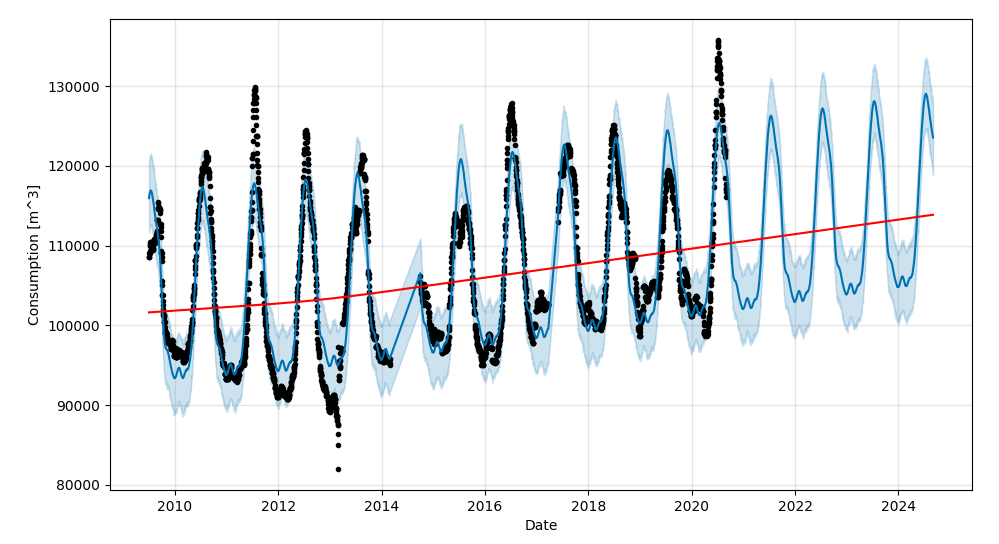}
    \caption{A forecast produced by the Prophet model trained on all available daily consumption estimates. The Prophet model's predictions are represented in blue, with error margins shown in light blue. Ground truth consumption values are represented by black points. The red line indicates the trend component of the model.}
    \label{fig:final-prophet-4year-forecast} 
\end{figure}

Returning to the original stakeholder requirements for this study, recall that a forecast of four years was to be utilized for budgetary projections. The Prophet model was trained on the entirety of the available preprocessed data and a forecast was obtained for $4$ years into the future. Figure~\ref{fig:final-prophet-4year-forecast} depicts the 4-year forecast for water consumption in the London region. Note that the trend was fixed after the first $80\%$ of the training data (as it was for all other trials), ensuring that recent irregularities (such as COVID-19) did not bias the trend leading into the forecast. This is an especially important practice, since Prophet extends the most recent trend into its forecast, calculating no further changepoints. The forecast is then simply the continuation of the trend into the future, with seasonal and holiday effects applied. In production, new Prophet models will be trained when data becomes available to maximize forecast accuracy. 

\section{Discussion}
\label{sec:discussion}

The water demand forecasting Prophet model described in section~\ref{subsec:final-model} is able to produce a reasonable water demand forecast. It was especially convenient that Prophet was able to handle missing data without introducing bias into forecasts, enabling us to utilize the entire available history of water consumption data. A quick glance at the daily estimates for citywide water consumption (see Figure~\ref{fig:preprocessed-data}) provides enough information to understand why Prophet was a sound choice for the problem at hand --- for example, there is a strong, repetitive seasonal periodicity observable in the data. Peaks are visible in the summer months and troughs appear in the winter months, which are respectively the warmest and coldest seasons in London. \textcite{Liu2020} also found this to be the case when studying water consumption patterns for Edmonton, Canada.

\begin{figure}
    \centering
    \includegraphics[width=\textwidth]{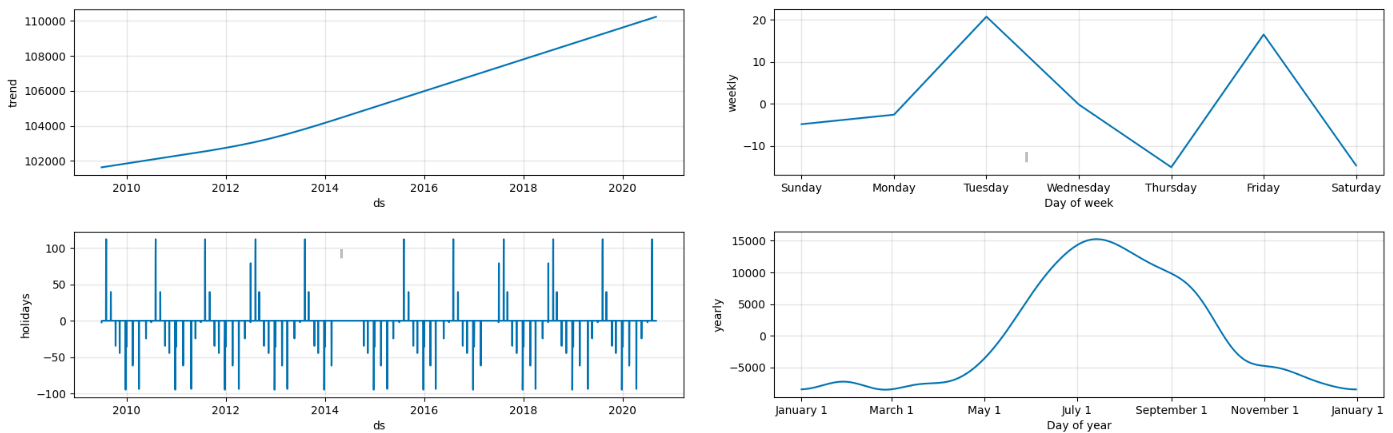}
    \caption{The components of the final Prophet model. Values on the y axes are expressed in cubic meters. Trend and holiday components, given for all dates in the training set (labelled "ds" on the x-axis), are presented in the top two graphs. The bottom two graphs display one period of the fitted yearly and weekly periodic functions.}
    \label{fig:final-prophet-components} 
\end{figure}

It was stated in section~\ref{subsec:model-selection}, Prophet is inherently interpretable. Recall from section~\ref{subsec:candidate-models} that Prophet predictions are a composition of linear trend, periodic functions for seasonal effects, and holiday effects. Further, holiday effects sprout from regionally specific data, which Prophet enables the analyst to implement as accessory information to improve forecasting. Prophet's forecasts can be decomposed, facilitating isolation of each component. Figure~\ref{fig:final-prophet-components} shows the four components of the final Prophet model: linear trend, yearly seasonality, weekly seasonality, and holiday effects. We will consider each of the components individually. First, the trend component indicates an overall growth in water consumption in London. Since the "changepoint prior scale" hyperparameter was set to an extremely low value, the number of changepoints in the trend has been minimized. The overarching growth in water demand over the last decade is consistent with London's continuing population growth, which has accelerated over recent years~\cite{StatisticsCanada2020}. Second, the holidays component indicates that water consumption is slightly lower on winter holidays and slightly higher on summer holidays. While this is interesting due to its congruence with annual seasonal effects, the effects are hardly noticeable, with no holiday parameter exceeding \SI{150}{m^3}. Similarly, the weekly component appears to imply that the day of the week has little bearing on Prophet's forecast. Such a result is unsurprising, considering the assumption that was made during data preprocessing that a client's daily consumption was uniform over the billing period, which is typically around a month in duration. In reality, weekly and holiday effects are likely more pronounced, since water consumption habits especially between Residential and Commercial rate classes, vary significantly between days of the week. Finally, yearly seasonality appears to contribute strongly to the model's prediction, varying by at least $20\%$ of the trend component. As discussed previously, the peak in summer months and dip in winter months correspond to the temperatures experienced in London, Canada throughout the year. Multiple activities may be contributing to this seasonal difference in consumption. Residential water use may spike in the summer due to increased frequency of citizens performing outdoor tasks requiring water, such as pool filling and lawn/plant watering. Commercial and industrial cooling processes that utilize water could also make up a substantial fraction of the elevated demand seen during hotter months. The decomposition of the Prophet model is an informative and highly valuable tool that will enable analysts to better understand long-term water demand forecasts.

Although this study achieved its goal of finding an accurate model for forecasting water demand in London, there remain opportunities for further study. First, multivariate models could be explored to discover whether other aspects of the water demand system have an impact on water usage forecasts. Customer attributes such as land area and meter type are examples of other possible predictors. Second, it may be worthwhile to study whether Prophet can accurately model water consumption for subsets of the population divided by their rate class. In London, all clients are designated as belonging to one of the following rate classes: residential, commercial, industrial, and institutional. Clients pay for water usage based on the rate assigned to their rate class. Models for each rate class would provide more detailed long-term estimates of municipal water revenue, and entail projections for revenue should the water rates be changed for a subset of the rate classes. Lastly, an investigation into the effect of climate severity on municipal water consumption and its implications for long-term forecasting would be of great interest.

\section{Summary}
\label{sec:summary}

Citywide water demand forecasting for London, Canada was posed as a univariate time series forecasting problem. Four models were trained, including OLS linear regression, RNN, CNN and Prophet models. Based on a 5-fold rolling origin cross validation, Prophet was discovered to minimize the primary metric of interest, achieving a MAPE of $2.51\%$ averaged over cross validation. Prophet was also able to perform forecasting despite missing segments of data. Furthermore, Prophet is inherently interpretable, the model itself being the sum of interpretable components for linear trend, seasonality, and holiday effects. A non-exhaustive survey of insights gleaned from Prophet's forecast and decomposition regarding water consumption were discussed. Avenues for subsequent study were also suggested. Finally, the methods described in this paper may be translatable to other municipalities; as a result, the implementation has been made public via a GitHub repository.

\printbibliography[title={References}]

%
%

\end{document}